\pdfoutput=1
\documentclass[letterpaper]{article}

   \usepackage{aaai24} 
   \usepackage{times} 
   \usepackage{helvet} 
   \usepackage{courier} 
   \usepackage[hyphens]{url} 
   \usepackage{graphicx} 
   \usepackage{graphicx}  
   \usepackage[sort]{natbib}  
   \usepackage{caption}  
\usepackage{listings}

\usepackage[utf8]{inputenc} 
\usepackage[T1]{fontenc}    
\usepackage[dvipsnames]{xcolor}

\usepackage{amsmath}
\usepackage{amssymb}
\usepackage{amsthm}
\usepackage{dsfont}
\usepackage{latexsym}
\usepackage{stackrel}
\usepackage{mleftright}

\usepackage{algorithm}
\usepackage{algpseudocode}

\usepackage[dvipsnames]{xcolor}
\usepackage[
    pagebackref=true,
    colorlinks=true,
    allcolors=MidnightBlue,
    ]{hyperref}
\usepackage{booktabs}       
\usepackage{nicefrac}       
\usepackage{multirow}
\usepackage{xspace}
\usepackage{placeins}
\usepackage[nameinlink,noabbrev]{cleveref}
\usepackage{dblfloatfix}

\usepackage[page]{appendix} 

\usepackage{siunitx}
\usepackage{lipsum}
\usepackage{xifthen}
\usepackage{tipa}

\usepackage{todonotes}

\DeclareCaptionStyle{ruled}{labelfont=normalfont,labelsep=colon,strut=off} 
\def\copyright@on{F}

\newcommand{\prob}[1]{\ensuremath{\mathop{{}\mathbb{P}}\mleft(#1\mright)}}

\newcommand{\set}[1]{\ensuremath{\left\{#1\right\}}}
\newcommand{\st}{\mathrel{}\middle|\mathrel{}}
\newcommand{\func}[2]{\ensuremath{\mathop{{}#1}\mleft(#2\mright)}}

\newcommand{\tipi}{\ensuremath{\tilde \pi}}

\newcommand{\pad}{\ensuremath{\textit{\textschwa}}}

\newtheoremstyle{thm}%
{1.em}%
{.5em}%
{\itshape}
{}
{\bfseries}
{}
{\newline}
{}

\theoremstyle{thm}

\newtheorem{theorem}{Theorem}
\newtheorem{proposition}[theorem]{Proposition}
\newtheorem{lemma}[theorem]{Lemma}

\renewcommand*{\backref}[1]{}
\renewcommand*{\backrefalt}[4]{%
\ifcase #1 %
  \relax
\or
  #2
\else
  #2
\fi
}

\title{Conformal Autoregressive Generation:\\Beam Search with Coverage Guarantees}
\author{
    Nicolas Deutschmann,
    Marvin Alberts,
    Mar\'ia Rodr\'iguez Mart\'inez
}
\affiliations{
    IBM Research\\
    \texttt{deu@zurich.ibm.com, marvin.alberts@ibm.com, mrm@zurich.ibm.com}
}

\begin{document}
\maketitle
\begin{abstract}
We introduce two new extensions to the beam search algorithm based on conformal predictions (CP) to produce sets of sequences with theoretical coverage guarantees. 
The first method is very simple and proposes dynamically-sized subsets of beam search results but, unlike typical CP procedures, has an upper bound on the achievable guarantee depending on a \textit{post-hoc} calibration measure. 
Our second algorithm introduces the conformal set prediction procedure as part of the decoding process, producing a variable beam width which adapts to the current uncertainty. 
While more complex, this procedure can achieve coverage guarantees selected \textit{a priori}. We provide marginal coverage bounds for each method, and evaluate them empirically on a selection of tasks drawing from natural language processing and chemistry.
\end{abstract}

\section{Introduction}
\label{sec:intro}
Autoregressive sequence models (ARSM) are probabilistic models describing distributions of sequence data, among which transformer-based large language models (LLMs)~\citep{vaswaniAttentionAllYou2023} have emerged as highly powerful tools for natural language processing (NLP) and generation (NLG). Beyond NLP, application of deep ARSMs in the natural sciences have generated impactful advances, such as in chemistry~\citep{schwallerMolecularTransformerModel2019,bornRegressionTransformerEnables2023} and molecular biology~\citep{rivesBiologicalStructureFunction2021}.

Despite their success, generating reliable predictions with ARSMs and quantifying their uncertainty combines the general challenges of trustworthy predictions in deep learning~\citep{guoCalibrationModernNeural2017a} with issues arising from the combinatorially large prediction space and the iterative processes used for sequence generation~\citep{gleaveUncertaintyEstimationLanguage2022,xiaoUncertaintyQuantificationPretrained2022,malininUncertaintyEstimationAutoregressive2021}. LLMs also present the unique further dimension of linguistic calibration~\citep{yinLargeLanguageModels2023,kadavathLanguageModelsMostly2022,mielkeReducingConversationalAgents2022}. Nevertheless, the high-profile of LLMs and the potential scientific value of ARSMs on other data modalities make the improvement of reliable uncertainty estimation methods an important and active area of research~\citep{kuhnSemanticUncertaintyLinguistic2023,jiangHowCanWe2021,schusterConfidentAdaptiveLanguage2022,linGeneratingConfidenceUncertainty2023,gleaveUncertaintyEstimationLanguage2022,fomichevaUnsupervisedQualityEstimation2020,fannjiangConformalPredictionDesign2022}.

A particularly interesting framework to infuse a notion of uncertainty or error control in predictions is Conformal Predictions (CP), which produces dynamically-sized \textit{sets} of predictions with finite-sample, distribution-free coverage guarantees~\citep{vovkAlgorithmicLearningRandom2005}. Beyond the formal guarantees, well-calibrated conformal sets are expected to reflect uncertainty through their size.

Generative tasks, where predicting is significantly harder than verifying a prediction seem like a perfect use case for the guarantees of CP. However, its original formulation relies on exhaustive searches that are untractable for sequences. Until recent methods relying on sampling~\citep{quachConformalLanguageModeling2023} or human-assisted pruning~\citep{renRobotsThatAsk2023}, solutions only existed at the token level~\citep{ravfogelConformalNucleusSampling2023,deyConformalPredictionText2022}. 

While sampling is particularly well-suited to NLG~\citep{holtzmanCuriousCaseNeural2020}, other 
applications that search correct or optimal solutions are better served by maximization-oriented decoding algorithms such as greedy or beam searches. Our work proposes to fill the current absence of such conformal decoding algorithms by introducing two new greedy methods for which we prove coverage guarantees and supported by emprical evaluation.

\section{Related Work}
\label{sec:related_work}
\paragraph{Conformal Predictions and Tolerance Regions}
Our work builds on top of the extensive literature developing CP, a set of tools for uncertainty estimation and trustworthy prediction first introduced by~\citet{vovkMachineLearningApplicationsAlgorithmic1999,vovkAlgorithmicLearningRandom2005,vovkConditionalValidityInductive2012}. This foundational work builds on the much earlier distribution-free tolerance regions developed by~\citet{wilksDeterminationSampleSizes1941,waldExtensionWilksMethod1943,tukeyNonParametricEstimationII1947} whose iterative construction is the basis of our step-by-step calibration method. In recent years, much activity has been spent developing CP to tackle modern machine learning~\citep{leiDistributionFreePrediction2012,angelopoulosUncertaintySetsImage2022}. Although not directly used in this paper, conformal risk control is of particular interest for potential extensions of our work~\citep{angelopoulosConformalRiskControl2023,angelopoulosLearnThenTest2022}.
\Citet{angelopoulosGentleIntroductionConformal2022} provide an excellent review of foundations and recent developments.

\paragraph{Uncertainty Estimation for Sequence models}
Given the high visibility of large language models and their unique position as powerful ML models directly in the hands of the general public, there is a strong interest in developing methods to ensure that their predictions are trustworthy and reliable, and to quantify the degree of uncertainty of model outputs. Current models tend to formulate wrong statements with an authoritative tone~\citep{yinLargeLanguageModels2023,kadavathLanguageModelsMostly2022,mielkeReducingConversationalAgents2022}, and measures of confidence built from model internals also tend to under-estimate errors~\citep{xiaoUncertaintyQuantificationPretrained2022,desaiCalibrationPretrainedTransformers2020,vasconcelosGenerationProbabilitiesAre2023}, motivating the development of improved uncertainty measures and calibration methods~\citep{schusterConfidentAdaptiveLanguage2022,kuhnSemanticUncertaintyLinguistic2023,malininUncertaintyEstimationAutoregressive2021,jiangHowCanWe2021}.

\paragraph{Conformal Predictions for Sequence Models}
The complexity of sequence generation make applications of CP non-trivial. As a result, early work on conformal prediction for ARSMs focused on applications such as few-shot classification~\citep{fischFewshotConformalPrediction2021,fischEfficientConformalPrediction2021} or single-token predictions~\citep{ravfogelConformalNucleusSampling2023,desaiCalibrationPretrainedTransformers2020}. Recently however, two exciting new methods proposed by~\citet{renRobotsThatAsk2023} and~\citet{quachConformalLanguageModeling2023} improved the state of the art significantly with solutions to handle general sequences. From a generation viewpoint, the work of~\citet{renRobotsThatAsk2023} focusing on task planning is the closest to ours: their goal is to generate high-scoring sequences of actions. A key difference is that our procedure generates sets of sequences instead of a product of sets of tokens, which reflects the domain disparity: robots must actually \textit{take} a single actions while we can generate multiple sequences. We also use step-wise thresholds to reflect the shifts in probability distributions through the decoding process, which is less relevant for planning as the environment is determined by more than the actions taken. Conversely, the decoding procedure of~\citet{quachConformalLanguageModeling2023} relies on sampling and semantic criteria for sequence-level acceptance and rejection based on Learn-then-Test~\citep{angelopoulosLearnThenTest2022} and is therefore quite different from our token-per-token approach. Theirs is thus better suited for the complex but fuzzy criteria of many NLG tasks. Both methods produce sets of full sequences with coverage guarantees, but our predictions are targeted toward controlling type-I errors on one-to-one -- or very tight -- input-sequence relationships while theirs aim at capturing a plurality of valid answers.
\section{Background}
\label{sec:background}
\subsection{Autoregressive sequence generation}
\label{sec:autoregressive}
Autoregressive models are a class of probabilistic models that describe a conditional probability mass function (PMF) $\func{\pi}{S|X}$ over a set of sequences $S\in A^*\omega$, \textit{i.e.} finite sequences over an alphabet $A$, terminated by a disjoint token $\omega$, given a condition $X$. The overall PMF $\pi$ is obtained as a decomposition into next-token probabilities $\tipi$ over the extended alphabet $\bar{A}=A \cup \set{\omega}$ conditioned on the right-truncated subsequences of $S=s_1\dots S_N\omega$:
\begin{equation}
    \pi(S) =\tilde\pi\left( \omega | s_1\dots s_N \right) \prod_{k=1}^{N} \tilde\pi\left(s_k | s_1\dots s_{k-1} \right).
\end{equation}
Deep learning models popular in NLP and beyond such as transformers~\citep{vaswaniAttentionAllYou2023} and iterations of recurrent neural networks~\citep{hochreiterLongShortTermMemory1997} provide learnable functions $\tipi$ that have had resounding success in recent years.

The access to $\tilde \pi$ is sufficient to tractably sample from $\pi$: starting from an empty sequence, next tokens are sampled from $\tilde \pi$ until $\omega$ is reached. However, obtaining a \textit{prediction} $S(X)$ in the typical machine learning sense,
\begin{equation}
    S(X) = \stackrel[S' \in A^*\omega]{}{\text{argmax}}\pi\left(S'|X\right),
    \label{eq:prediction}
\end{equation}
is not a solved problem and is usually addressed with heuristic methods. The simplest such approach is greedy search where the candidate sequence is extended with the most-probable next token. Greedy approaches fail if an early high-probability token only leads to lower-probability sequences and are therefore often extended using beam-search (BS) approaches where $b$ candidate sequences are considered and greedily extended, keeping the $b$ highest-scoring extensions at the next step.

The lack of theoretical grounding for these algorithms has two implications for generation-as-prediction tasks:
\begin{itemize}
    \item there is no rigorous way to produce \textit{the} prediction of the model for a given condition $X$ (\cref{eq:prediction}),
    \item given a heuristic proposal, the \textit{correctness} of the prediction cannot easily be studied from a theoretical point of view.
\end{itemize}
In this work, we address exclusively the second issue: given a pair $(X,S)$ sampled from the true distribution $p$, we leverage conformal predictions to provide a greedy set of candidate sequences from $\pi$ with a guarantee on the probability that $S$ is included.

\subsection{Conformal predictions}\label{sec:confpred}
Conformal predictions are a framework to produce sets of predictions $C(X)$ from a predictive model and a measure of confidence and uncertainty $s(X,y)$.

Our method falls under the umbrella of Split CP~\citep{papadopoulosInductiveConfidenceMachines2002}, where we consider a pre-trained predictive model, a prediction confidence score function $s(X,y)$ and exchangeable data $\set{(X_i,y_i)}_{i=1,\dots N+1}$. Considering the set $\mathcal{C}_N$ of the first $N$ samples as calibration data and choosing a risk level $\alpha$, we define a threshold $t_\alpha$ for scores as the $k_\alpha^{(N)}$-th smallest calibration score, where
\begin{equation}
    k_\alpha^{(N)} = \lfloor \alpha (N+1) \rfloor.
    \label{eq:cp_thresh}
\end{equation}
Defining a prediction set $\func{C}{X_{N+1}}$ for the $(N+1)$-th sample as 
\begin{equation}
    \func{C}{X_{N+1}} = \set{y'\in \Omega_Y | \func{s}{X,y'}\geq t_\alpha},
    \label{eq:cp_predset}
\end{equation}
where $\Omega_Y$ is the space of possible labels $y$.

Marginally over the calibration set, the following inequality holds true:
\begin{equation}
\prob{y_{N+1}\in \func{C}{X_{N+1}}} \geq 1-\alpha,
\label{eq:cp_guaratee}
\end{equation}

\section{Methods}
\label{sec:theory}
We present two methods for obtaining greedy conformal prediction sets from autoregressive models: the first is based on proposing a dynamically-sized subset of the usual beam search results, while the second uses CP to choose a beam size dynamically at each step.

\subsection{Conformal Predictions from Beam Search Results}
\label{sec:fixed_beam_theory}
Let us start with the simplest approach, where we rely on the standard beam search algorithm.

Given a pair $(X,S)$ sampled from $P(S,X)$, we wish to produce a set $C(X)$ of candidate sequences with some form of guarantee on $\prob{S\in C(X)}$. The combinatorially-large sequence space makes a direct application of~\cref{eq:cp_predset} untractable. An alluring alternative would be to start by performing a beam search, in order to obtain a reasonably-sized set of proposals $\func{\beta}{X}$, and to then predict $C(X)\subset B(X)$. This however, only works if we can provide a coverage guarantee on $\func{\beta}{X}$, which we do below.

The procedure itself relies on group-conditional conformal predictions~\citep{vovkConditionalValidityInductive2012}. Using the notation from \cref{sec:confpred}, we define the in-beam subgroup of the calibration data $\mathcal{C}_N$:
\begin{equation}
     \mathcal{C}_{\beta}=\set{(X_i,S_i) \in \mathcal{C}_N | S_i\in \func{\beta}{X_i}}.
\end{equation}
Defining $N_\beta = \left|\mathcal{C}_\beta\right|$, we perform split-CP calibration on this subgroup, with a confidence score $\func{s}{X_i,S_i}$ such as the ARMS probability $\func{\pi}{S_i|X_i}$. We thus obtain a threshold $t_\alpha^{(N_\beta)}$ as the $k_\alpha^{(N_\beta)}$-th smallest score as defined in \cref{sec:confpred}.

At inference time, given a test sample $(X, S)$, we define the prediction set
\begin{equation}
    \func{C_{\alpha|\beta}}{X} = \set{S'\in \func{\beta}{X} | \func{\pi}{S'|X}\geq t_{\alpha}}.
    \label{eq:fixed_beam_set}
\end{equation}

On this prediction set, we provide the following guarantee:


\begin{proposition}
With probability at least $1-\delta$,
\begin{multline}
    \prob{S\in \func{C_{\alpha|\beta}}{X}} \geq \\
     (1-\alpha)\func{B}{\delta; N_\beta, N+1-N_\beta}.
    \label{eq:marginal_beam_cov}
\end{multline}
where $B(a,b)$ is a beta distribution and $B(\delta; a,b)$ is its $\delta$-quantile.
\label{thm:marginal_beam_cov}
\end{proposition}

This result follows from the conditional decomposition of the coverage probability:
\begin{multline}
    \prob{S\in \func{C_{\alpha|\beta}}{X}} = \\
    \prob{S\in \func{C_{\alpha|\beta}}{X} | S\in \func{\beta}{X}} \times \prob{S\in\func{\beta}{X}}.
\end{multline}
The conditional conformal procedure described above guarantees that the first term is at least $1-\alpha$, following from~\citet{vovkConditionalValidityInductive2012}. Unlike conformal guarantees, the second bound is obtained from the observation of $N_\beta$, which follows a binomial distribution. While we cannot decide its success probability, we can use confidence intervals from~\citet{clopperuseconfidencefiducial1934} to provide a bound with risk $\delta$.

\subsection{Dynamic Conformal Beam Search}
\label{sec:dynamic_conformal_beam}
In this section, we present our second method to provide conformal prediction sets for ARSM. This approach relies on choosing a dynamic beam size at each decoding set based on a CP threshold, which permits a pre-determined guarantee.

\subsubsection{Calibration Algorithm}
We consider $N_0+1$ exchangeable pairs $\set{(X_i,S_i)}$ and a family of conformal scores $\sigma_l$ that can be evaluated on length-$l$ sequences. Selecting the first $N_0$ samples as $\mathcal{C}^{(0)}_{N_0}$, we specify a per-step confidence level $1-\alpha$ calibrate iteratively as follows: at the $l$-th step,
\begin{enumerate}
    \item Define $k_\alpha^{(l)}$ = $\lfloor(N_{l-1}+1)\alpha\rfloor$.
    \item Order the calibration set by increasing length-$l$ scores $\func{\sigma_l}{X_1,S_{1|l}}\geq \dots \func{\sigma_l}{X_{N_{l-1}},S_{N_{l-1}|l}} $, where $S_{i|l}$ is the length-$l$ truncation of $S_{i}$.
    \item Define $t_{\alpha,N_{l-1}}^{(l)}=\func{\sigma_l}{X_{k_\alpha^{(l)}},S_{k_\alpha^{(l)}|l}}$.
    \item Set $N_{l} = N_{l-1} - k_\alpha^{(l)}$, $\mathcal{C}^{l}_{N_{l}} = \set{(X_i,S_i)}_{k_\alpha^{(l)}<i}$.
\end{enumerate}
The iteration can in-principle be continued infinitely if $\sigma_l$ is based on the model $\tipi$ and we extend $\tipi$ defined in~\cref{sec:autoregressive} by specifying that $\tipi$ predicts a padding token $\pad\in A$ with probability 1 after the terminating token $\omega$.

We define the set of acceptable length-$L$ sequences as $\func{\Omega_L}{A,\omega, \pad}=\left[\bar{A}^* \cdot \omega^? \cdot \pad^* \right]_L$, \textit{i.e.} sequences of the form $a_1\dots a_k \omega^? \pad \dots \pad$ with $0\leq k \leq L$, an optional terminating omega $\omega$ and $L-k-1$ padding tokens. 

\subsubsection{Inference Algorithm}
At inference, we consider the left-out sample $X_{N_{0}+1},S_{N_0+1}$, dropping the index for brevity. The first decoding step is a standard conformal prediction on length-one sequences:
\begin{equation}
    \func{C_\alpha^{(1)}}{X} = \set{S'_{|1} \in A \st \sigma_1{X,S'_{|1}}\geq t_{\alpha,N_0}^{(1)}}.
\end{equation}
Proceeding iteratively until all sequences in $ \func{C_\alpha^{(l)}}{X}$ terminate or a maximum length $L^\dagger$ is reached, we define the next conformal beam as all continuations in the previous beam that pass the next threshold
\begin{equation}
    \def\arraystretch{2}
    \func{C_\alpha^{(l+1)}}{X} = \set{
     S'_{|l}\, a \left|
    \begin{array}{l}
    a\in\bar{A},\, S'_{|l} \in \func{C_\alpha^{(l)}}{X},\\
    \func{\sigma_{l+1}}{X,S'_{|l}a} \geq t_{\alpha,N_{l}}^{(l+1)}
    \end{array}\right.
    }.
    \label{eq:dynamic_prediction_set}
\end{equation}
This approach mirrors the traditional beam-search algorithm in that it keeps a set of proposals at each decoding step and generates a set of high-scoring continuations of the current proposal for the next step.

\subsubsection{Guarantees}
The iterative subselection procedure that defines the subsequent conformal thresholds $t_{\alpha,N_{l}}^{(l)}$ defines a multivariate tolerance region as defined in~\citet{waldExtensionWilksMethod1943} and \citet{tukeyNonParametricEstimationII1947} which provide distribution-free coverage guarantees. Indeed, our thresholds correspond to iteratively pruning $\mathcal{C}^{(0)}_{N_0}$ by removing the $k_\alpha^{(l)}$-lowest scoring element based on a scoring function applied to the samples. Our set of thresholds defines a confidence region $\func{R}{\mathcal{C}^{(0)}_{N_0}, \alpha, L}$ as 
\begin{multline}
 \func{R}{\mathcal{C}^{(0)}_{N_0}, \alpha,L} = \bigg\{S' \in \func{\Omega_L}{A,\omega,\pad}\, \mid \forall 0\leq l < L\\ \func{\sigma_{l+1}}{X,S'_{|l}a} \geq t_{\alpha,N_{l}}^{(l+1)}\bigg\}.
\end{multline}
Its conditional coverage probability given a calibration set $\mathcal{C}^{(0)}_{N_0}$ is a random variable, which, strikingly, depends only on $\alpha$ and $N_{0}$:
\begin{multline}
    \prob{(X,S)\in \func{R}{\mathcal{C}^{(0)}_{N_0}, \alpha,L}\Big| \mathcal{C}^{(0)}_{N_0}} \\ \sim \func{B}{N_0+1-\sum_l k_\alpha^{(l)}, \sum_l k_\alpha^{(l)} }.
  \end{multline}

Marginalizing over $\mathcal{C}^{(0)}_{N_0}$, we obtain
\begin{lemma}
\begin{align}
    \prob{\set{\func{\sigma_{l+1}}{X,S} \geq t_{\alpha,N_{l}}^{(l+1)}}_{l\leq L}} &
    = 1 - \sum_{l=1}^L \frac{k_\alpha^{(l)}}{N_0+1}\\
    & \geq (1 - \alpha)^L.
    \label{eq:coverage_region_guarantee}
\end{align}
\end{lemma}
Importantly, notice that since the threshold at step $l$ is applied exhaustively on all sequences that passed the previous threshold, the prediction set defined in~\cref{eq:dynamic_prediction_set} corresponds to all sequences $S'$ that verify the condition of~\cref{eq:coverage_region_guarantee} given $X$:
\begin{multline}
    \func{C_\alpha^{(L)}}{X} = \bigg\{ S'\in \func{\Omega_L}{A,\omega,\pad} \,\Big|\\
    \set{\func{\sigma_{l+1}}{X,S} \geq t_{\alpha,N_{l}}^{(l+1)}}_{l\leq L}\bigg\}.
\end{multline}

Extending the decoding until a maximum length $L^\dagger$, we obtain the guarantee that for $(X,S)\sim P(X,S)$ exchangeable with the calibration set,
\begin{proposition}
\begin{equation}
    \mathop{{}\mathbb{P}}\bigg[ S \in \func{C_\alpha^{(L^\dagger)}}{X} \bigg] \geq (1 - \alpha)^{L^\dagger}.
\end{equation}
\label{thm:dynamic_beam_cov}
\end{proposition}

\section{Experimental results}
\label{sec:experimental}
\subsection{Datasets and Models}
We evaluate our conformal generation algorithms on two transformer models, each evaluated on a generative task that squarely fits in the category of predictions, rather than distribution modelling, in the sense that there is a single \textit{correct} sequence given an input $X$.

\subsubsection{Integer Additions}
We finetune an off-the-shelf \texttt{t5-base} sequence-to-sequence model from HuggingFace on a simple addition task: computing the sum of two integers with up to seven digits each. The problem is formulated as a sequence-to-sequence prediction task, matching arithmetic problems of the form "\texttt{1789+111=}" to results presented as a sequence of digits followed by an EOS token, "\texttt{1900$\omega$}". We sample 130k such additions which we split into 100k training examples and 30k held out samples. On the validation set, this model has a mean coverage of 96\% using 5-sequence beam search. This is a short-sequence task with a high performance model, making it a best-scenario test case for our dynamic conformal decoding algorithm.

\subsubsection{Chemical Reaction Product Prediction}
We train a \texttt{t5-small} from scratch to predict the product of chemical reactions, using SMILES strings~\citet{weiningerSMILESChemicalLanguage1988} to encode reagents and products. We use the USPTO-MIT dataset~\citep{jin_predicting_2017} and tokenization scheme by Schwaller et al.\ ~\citep{schwallerMolecularTransformerModel2019} for training and evaluation, holding out 30k samples with length lower than 50 for calibration and testing. On the validation set, this model has a mean coverage of 64\% using 5-sequence beam search. In contrast to the additions task, this task features much longer sequences and a lower-accuracy model, making it a more challenging testing ground.

\subsection{Conformal Beam Subsets}
We start by evaluating the success rate $1-\delta$ of the composite bound of \cref{thm:marginal_beam_cov}. To this end, we perform 1000 bootstrapped estimates, sampling twice 15k held-out samples for calibration and test. For each repetition, we measure $N_\beta$, evaluate the Clopper-Pearson guarantee and perform the conditional-in beam conformal calibration. On test data, we evaluate the conditional and global coverage and evaluate the success rate of the inequality in \cref{thm:marginal_beam_cov} over the repetitions. As we show in \cref{tab:beam_cov_test}, the bound does hold with sufficiently low risk. For all experiments, we use the length-normalized sequence probability under the model $\func{\pi}{S|X}/|S|$ as the conformal confidence score.

\begin{table*}[t]
    \renewcommand{\arraystretch}{1.2}
    \centering
    \begin{tabular}{cc|c|cccc}
    \hline
    \multirow{2}{*}{Task} & \multirow{2}{*}{Beam Size} & \multirow{2}{*}{Beam Cov.} & \multicolumn{4}{c}{$1-\alpha =0.95$} \\
    & & & Conditional Cov. & MAE & Global Bound & Global Cov. \\\hline
    
    \multirow{2}{*}{Additions (T5)}  
    & 10 & $0.979(9)$ & $0.949(2)$ & $0.73(2)$ & 0.910(1) & 0.912(2)\\
    & 5  & $0.961(1)$ & $0.950(2)$ & $0.45(1)$ & 0.911(1) & 0.913(2)\\\hline
    
    \multirow{2}{*}{Reactions (T5)}  
    & 10 & 0.699(2) & 0.949(2) & 4.48(7) & 0.658(2) & 0.663(3)\\
    & 5 & 0.641(3) & 0.949(3) & 1.86(4) & 0.602(3) & 0.608(4)\\\hline\hline
                                     
    \multirow{2}{*}{Task} & \multirow{2}{*}{Beam Size} & \multirow{2}{*}{Beam Cov.}  & \multicolumn{4}{c}{$1-\alpha =0.99$} \\
    & & & Conditional Cov. & MAE & Global Bound & Global Cov. \\\hline
    \multirow{2}{*}{Additions (T5)}  
    & 10 & $0.979(9)$ & $0.989(1)$ & $2.43(7)$ & 0.947(1) & 0.951(1)\\
    & 5  & $0.961(1)$ & $0.989(1)$ & $1.33(3)$ & 0.948(1) & 0.952(2)\\\hline
    \multirow{2}{*}{Reactions (T5)}  
    & 10 & 0.699(2) & 0.989(1) & 6.69(7) & 0.683(3) & 0.692(2) \\
    & 5 & 0.641(3) & 0.989(1) & 2.82(3) & 0.625(1) & 0.634(3)\\\hline
    
    \end{tabular} 
    \caption{We report the conditional coverage and mean absolute error between the prediction set size and the rank of the correct sequence in the beam on 1000 bootstrapped experiments. The uncertainty on the last significant digit is given as the standard deviation across the experiments.}
    \label{tab:beam_cov_test}
\end{table*}

\begin{figure*}[t]
    \centering
    \includegraphics[width=0.3\linewidth]{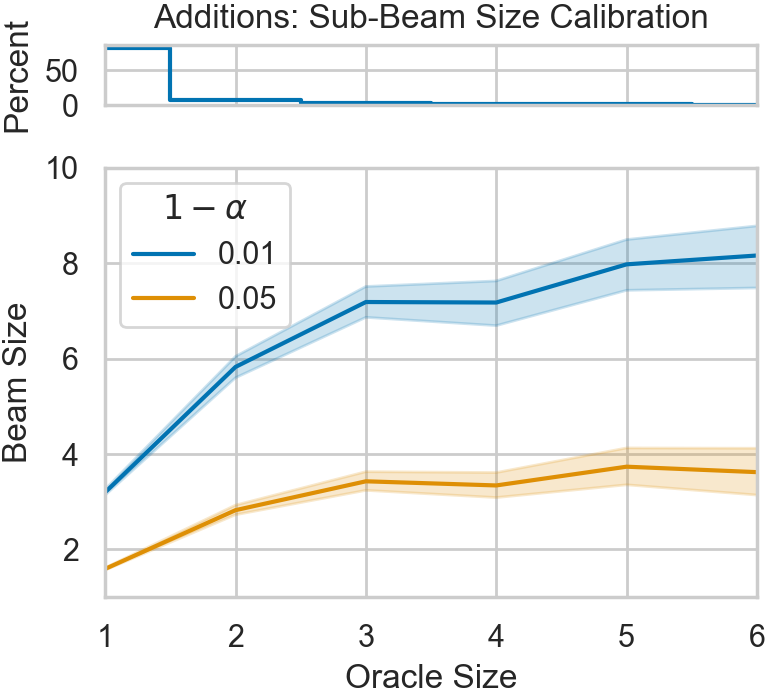}\quad
    \includegraphics[width=0.3\linewidth]{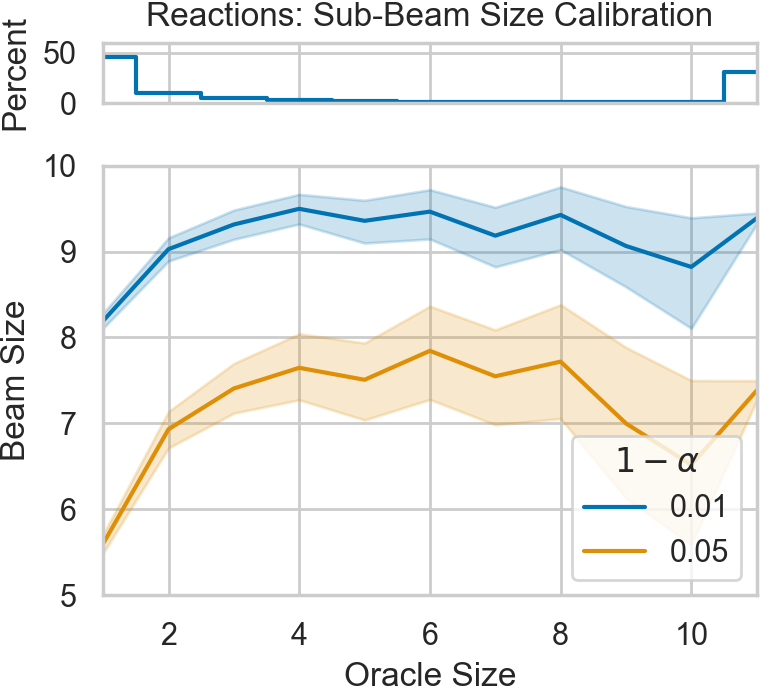}
    \caption{(Main) Conformal sub-beam set sizes plotted against oracle set sizes for the addition task (left) and the chemical reaction prediction task (right). (Top) Oracle set size distribution.}
    \label{fig:sub_beam_set_sizes}
\end{figure*}

While not guaranteed, a desirable feature of conformal prediction sets is that their size is informative of some notion of uncertainty. If predictions set sizes closely follow the rank of the correct sequence in the beam, ordered by scores, they can be used as a measure of the quality of the prediction. We measure this by considering the mean absolute error between the predicted set size and that of a perfect oracle with set sizes exactly matching the true sequence rank if it is in the beam, and predicting the entire beam if not. We also report this information in \cref{tab:beam_cov_test}. While the additions model provide tight prediction sets, the reaction prediction model seems poorly calibrated. As we nevertheless show in \cref{fig:sub_beam_set_sizes}, the model is indeed over-conservative for low-rank sequences but for both models set size is a useful predictor of low versus high rank when taking rank distribution into account.

\subsection{Dynamic Conformal Beams}
Let us now turn to our second procedure where we use conformal predictions to iteratively decode with guaranteed coverage at each step. For our benchmark tasks, we exploit a simplification of our dataset definitions: we know the maximum sequence length in advance, respectively 5 and 50 tokens for additions and chemical reactions. We use this knowledge to set the maxumum number of decoding steps to 5 and 50 and avoid discussing rare long sequences. We discuss how to handle unknown maximum lengths in \cref{sec:limitation}.

For each benchmark task, we run multiple independent calibration-and-inference experiments, randomly sampling 30\% of the held-out data for calibration and 1500 test sequences.
Metrics are measured for each repetition and averaged, and we report the standard deviation across the repetitions.
The smaller scale of the additions task allowed us to run more experiments and we therefore have 120 independent repetitions for step-wise guarantees levels $(1-\alpha)\in [0.99, 0.98, 0.95]$ yielding sequence-level marginal coverage guarantees of $[0.95, 0.90, 0.77]$.
Predicting chemical reaction products is much more computationally demanding owing to the longer sequences and and larger model, as well as a lower accuracy, leading to larger prediction sets.
As a result we limit ourselves to 50 repetitions for two token-wise confidence levels $(1-\alpha)\in [0.995, 0.99]$, which are more strict to enable coverage control after 50 decoding steps. These token-wise values define sequence-wise guarantees of $[0.78, 0.60]$, which frame the 5-beam validation accuracy of the model of $0.64$.

As for the fixed-beam procedure, we start by empirically checking our coverage guarantees. We report observed coverages in \cref{tab:global_dynamic_beam_metrics}, showing that the mean observed coverage indeed dominates the guarantee. For lower confidence levels, the bounds tend to be less tight, owing to the lower frequency of longer-than needed decoding procedures.

\begin{table*}[t]
    \renewcommand{\arraystretch}{1.2}
    \centering
\begin{tabular}{cc|llcc}
\toprule
Task & $1-\alpha$ & Coverage & Guarantee & $ \left|\beta (X)\right| $ & $\frac{\left|\beta (X)\right|}{\left|\beta_\text{O} (X)\right|}$ \\
\midrule
\multirow{3}{*}{Additions} & 0.990 & 0.9607(3) & 0.9509 & 10.4(4) & 6.64(2) \\
 & 0.980 & 0.9244(4) & 0.9039 & 4.82(1) & 3.30(1) \\
 & 0.950 & 0.8231(6) & 0.7737 & 1.756(3) & 1.476(2) \\\midrule

\multirow{2}{*}{Reactions} & 0.995 & 0.806(1) & 0.778 & 47.(5) & 25.(2) \\
                           & 0.990 & 0.693(1) & 0.605 & 4.13(1) & 3.08(1) \\
\bottomrule
\end{tabular}
    \caption{Sequence-level metrics for the dynamic beam decoding procedure on the additions and chemical reaction tasks measured across respectively 120 and 50 repetitions. For each dataset and tested confidence level, we report the mean coverage and compare it to the guarantee of \cref{thm:dynamic_beam_cov}, as well as the mean beam size $\mathbb{E}\left|\beta (X)\right|$ and the mean beam-size to oracle beam-size ratio $\mathbb{E}\frac{\left|\beta (X)\right|}{\left|\beta_O (X)\right|}$. For each metric, we report the average and the uncertainty on the last digit measured as the standard deviation of the mean estimator between parentheses.}
    \label{tab:global_dynamic_beam_metrics}
\end{table*}

\Cref{tab:global_dynamic_beam_metrics} also reports mean beam sizes at the end of decoding $|\beta(X)|$ and compares them to an optimal oracle that predicts a subset $\beta_O(X)$ of our dynamic beams such that the lowest-ranking sequence is the correct one. While it was appropriate to use MAE for the fixed beam size method, the dynamic version does not have a reference size to which we could compare and we therefore found ratios more informative. 

\begin{figure*}[t]
    \centering
    \includegraphics[width=0.3\linewidth]{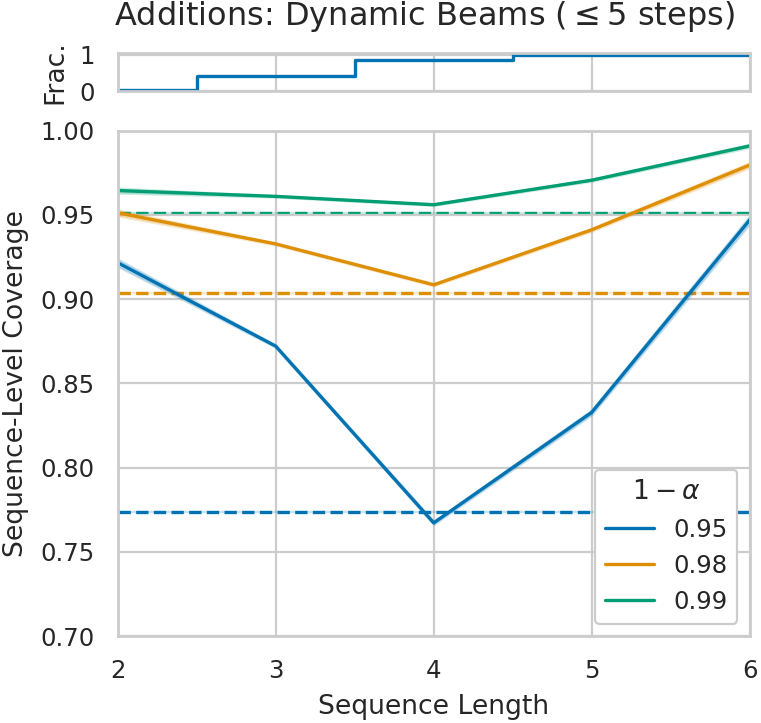}\quad
    \includegraphics[width=0.3\linewidth]{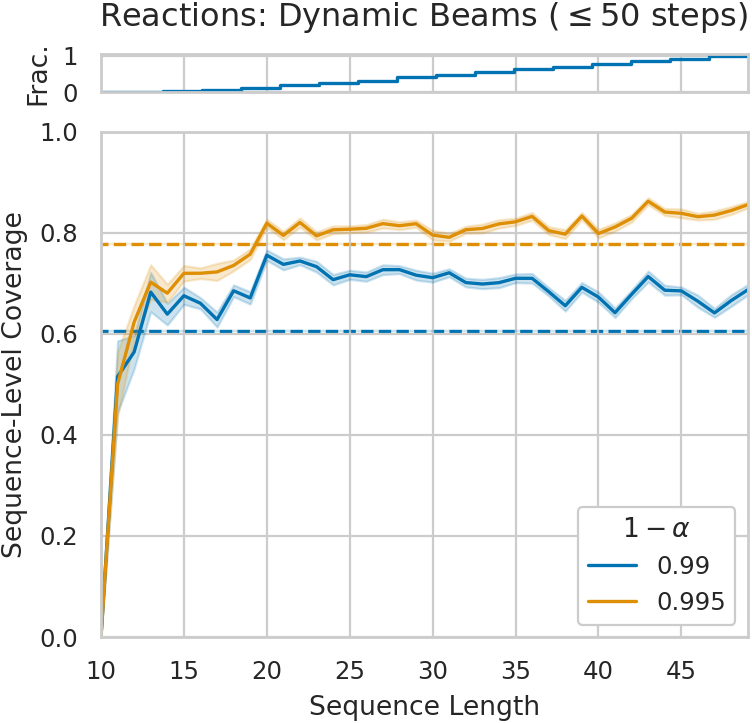}
    \caption{Per-sequence-length observe coverages for the additions (left) and reactions (right) tasks. Coverage guarantees are provided as dashed lines for each confidence level.}
    \label{fig:per_len_cov}
\end{figure*}

As show in \cref{fig:per_len_cov}, conditional coverage is not ensured per sequence length. While coverage is near-uniform across reaction product lengths, there is significant variation across addition sequence lengths.

\begin{figure*}[t]
    \centering
    \includegraphics[width=0.3\linewidth]{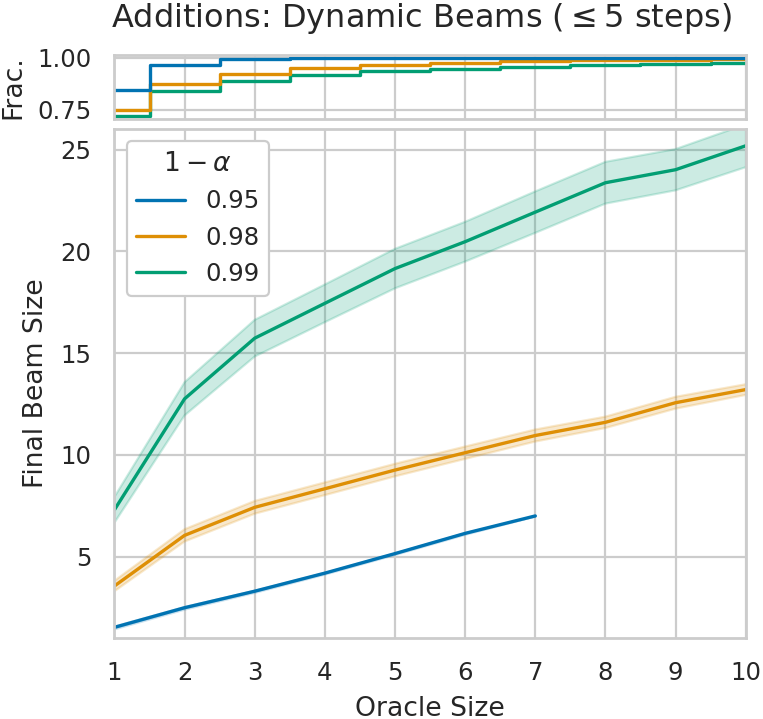}\quad
    \includegraphics[width=0.3\linewidth]{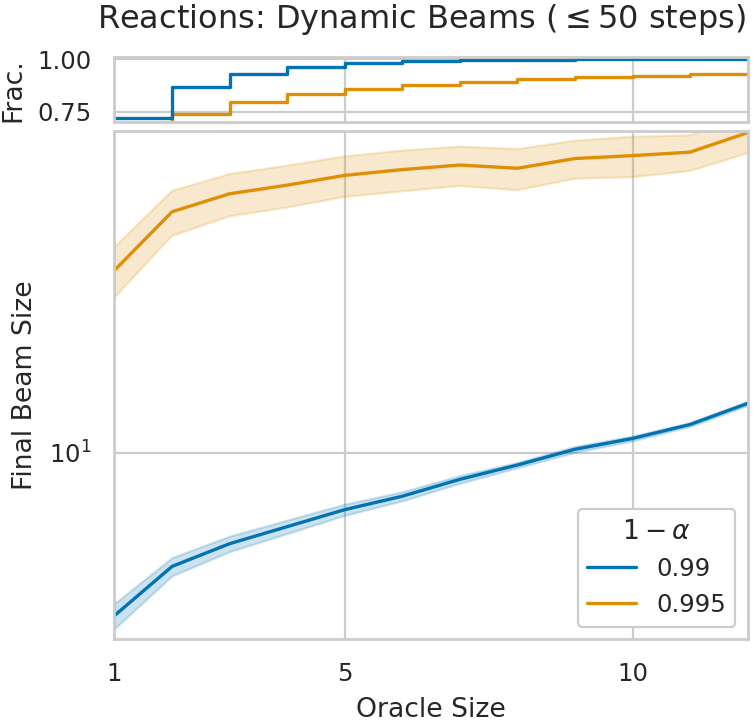}
    \caption{(Main) Dynamic conformal beam sizes plotted against oracle set sizes for the addition task (left) and the chemical reaction prediction task (right). (Top) Oracle set size distribution. A high correlation between oracle and beam sizes is indicative of good calibration.}
    \label{fig:dynamic_bs}
\end{figure*}

We characterize the adaptivity of the prediction set sizes by observing whether the beams are tight in the sense than larger beams are predicted are indicative of low-ranking true sequences, which we show to be the case in \cref{fig:dynamic_bs}. Beam size is indeed much more predictive of oracle size than for the conformal beam subset procedure, making dynamic conformal beam decoding preferable in terms of uncertainty quantification.

\begin{figure*}[t]
    \centering
    \includegraphics[width=0.3\linewidth]{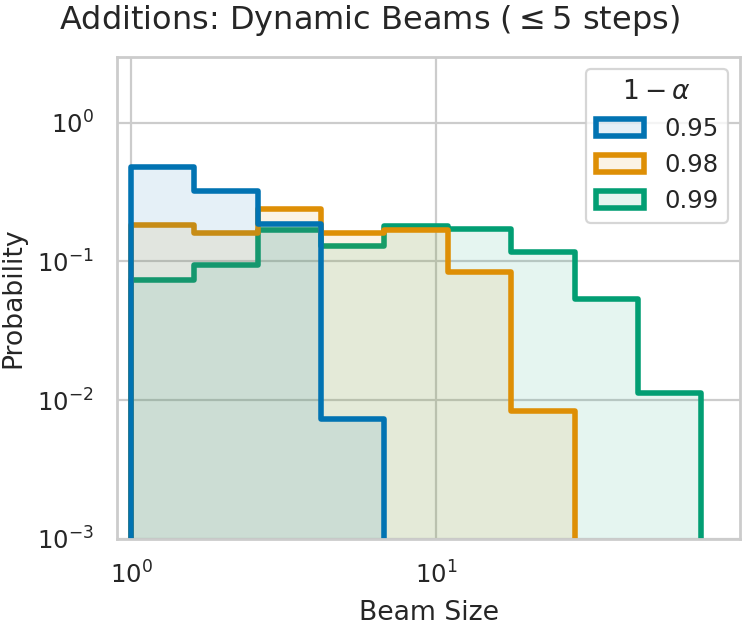}\quad
    \includegraphics[width=0.3\linewidth]{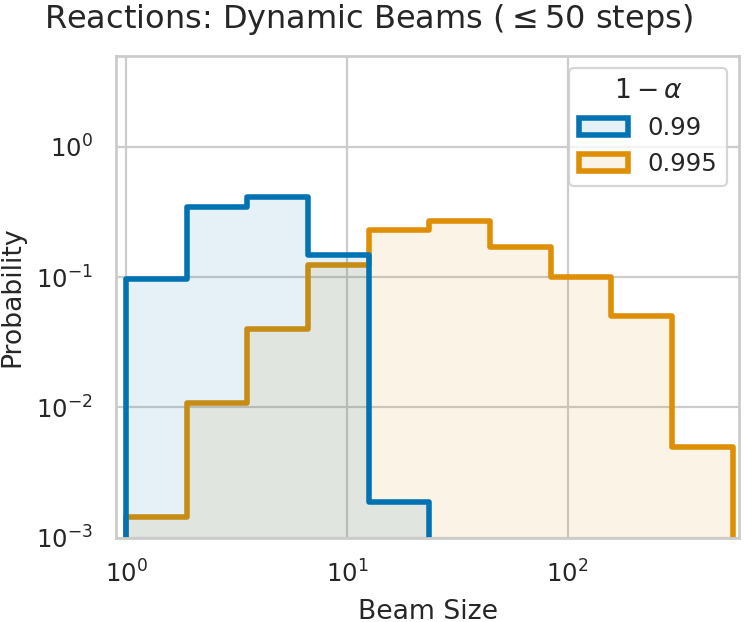}
    \caption{Final beam size distributions for the additions (left) and reactions (right) tasks.}
    \label{fig:dynamic_bs_hist}
\end{figure*}

One disadvantage of dynamic conformal beams is the absence of a bound on beam sizes, which can become large if the model doesn't have high performance, or sequences are long: either case requires a stringent per-token confidence level. This is indeed what we observe in the reaction prediction task with $\alpha=0.005$. As we show in \cref{fig:dynamic_bs_hist}, a significant fraction of beams contain more than 100 sequences, which is both computationally expensive and potentially undesirable, in the sense that many incorrect sequences are predicted.

\section{Limitations and Usage Recommendations}
\label{sec:limitation}
Our two methods have the potential to further the confident and uncertainty-aware use of generative sequence models. Nevertheless, they have clear limitations that must be made explicit for both practitioners and potential future development.
\begin{itemize}
    \item Most importantly, both methods rely on greedy search for high-scoring sequences. This is especially adapted for autoregressive models used as predictors over sequences, which is common in the sciences, but less so for language modelling except maybe in the case of reward optimization for reinforcement-learning-based models~\citep{gleaveUncertaintyEstimationLanguage2022}.
    \item As a hybrid conformal-estimation method, conformal beam subsets are not a prescriptive way to obtain a certain desired coverage, unlike traditional conformal methods. Only a fraction of the observed calibration coverage is attainable. Given that requiring too-high guarantees can lead to very large sets, this might be an acceptable compromise.
    \item Our dynamic conformal beam decoding procedure provides a per-decoding step guarantee which degrades exponentially with the number of steps executed. Choosing a low per-step risk might be acceptable for high-accuracy models but might become unwieldy for more difficult tasks.
    \item Dynamic beam guarantees also rely on a choice of maximum length which is an extra hyper-parameter to set and can affect performance: a too low cut-off will lead to failures for long sequences while the opposite choice degrades performance for shorter sequences.
\end{itemize}

Consequently, we make the following recommandations for practical use of our methods:
\paragraph{Method choice} For high-accuracy models on moderately sized models, using dynamic beams is probably the best-suited method. However, for tasks or models yielding reduced performance measured on fixed-size beams, one should consider whether the computational cost of very large dynamic beams is worth the benefits. As a conservative rule of thumb, we recommend to use fixed 5-sequence beam coverage as a reference for achievable dynamic performance with dynamic sizes up to the tens. Its applicability is of course model and task-dependent and a score distribution with a long tail should motivate the use of fixed beams instead.
\paragraph{Choice of maximum decoding length}
Our dynamic conformal beam decoding procedure relies on setting a maximum decoding length which affects the guarantee exponentially. In our experiments, we leveraged the knowledge of an actual maximum length in the dataset to set this limit but in many cases this knowledge is not accessible. Given that we cannot expect to evaluate uncertainties using a per-step procedure on sequences for which we have no or limited calibration data, setting this limit based on a high quantile of the observed length distribution is a safe choice. This limit should be set using other data than the calibration set to avoid breaking exchangeability, but the model training data is perfectly suitable for this purpose.
\paragraph{Making dynamic guarantees tighter with length-conditional calibration}
As shown in \cref{tab:global_dynamic_beam_metrics}, a significant fraction of sequences with decoding procedures shorter than the maximum can lead to less-than-tight guarantees. If the length distribution is spread, splitting sequences into groups of similar lengths and using the group-conditional calibration procedure introduced by~\citet{vovkConditionalValidityInductive2012} is well-advised. At decoding time, a two-step procedure can be used: first decoding with the minimum threshold across all groups until termination. At this point, a group can be assigned to the decoded beam and pruned based on the group-specific thresholds. Each length group has the coverage guarantee of its longest sequences.
\section{Conclusion}
\label{sec:conclusion}
The two methods we introduce in this paper further the applicability of split-conformal prediction sets with distribution-free guarantees to autoregressive sequence generation tasks. Our approach complements the recent work of~\citet{quachConformalLanguageModeling2023} based on sampling by using a greedy decoding approach which makes it better suited for predictive tasks. Extending our approach to proper language modelling would nevertheless be interesting, especially for models trained or fine-tuned with reinforcement learning~\citep{ouyangTrainingLanguageModels2022}. Another exciting further development would be to use our method to restrict the sampling options of \citet{quachConformalLanguageModeling2023}, which might improve both performance and reduce its rejection rate~\citep{holtzmanCuriousCaseNeural2020,cohenEmpiricalAnalysisBeam2019}.

\FloatBarrier

\section*{Acknowledgements}
We thank Jannis Born, Dimitrios Christofidellis, Mattia Rigotti and Alain Vaucher for helpful discussions and advice.
The work of ND was supported by the Swiss National Science Foundation Grant No. 192128.

\bibliography{biblio}

\appendix

\section{Experimental Details}

\subsection{Additions Task}
\subsubsection{Dataset}
The additions task is defined as a sequence-to-sequence prediction task where integer additions formulated as natural language are matched to the result of the addition. Questions are expressed as strings of the form \texttt{112+17=} and the result as \texttt{129$\omega$}, where $\omega$ is the end-of sequence token.

The dataset is generated with code to be open-sourced is composed of all one- and two-digit integers in the left and right positions, as well as randomly sampled additions with more digits according to the following procedure:
\begin{itemize}
    \item select two numbers of digits, $m$ and $n$.
    \item Sample $N=5000$ numbers of each size $x_{i}^{(m)}$ and $x_{i}^{(n)}$.
    \item Repeat $k=10$ times:
    \begin{itemize}
        \item permute the indices of $x_{i}^{(m)}$ and $x_{i}^{(n)}$,
        \item sample $l_{1i},l_{2i}$, $N$ random permutations of the pair $(m,n)$,
        \item generate addition problems-answer pairs for the pair of integers $x_{i}^{(l_{1i})},x_{i}^{(l_{2i})}$.
    \end{itemize}
\end{itemize}
We do this for the following pairs $(m,n)$:
\begin{align*}
&(3, 3),
(2, 4),
(3, 4),
(4, 4),
(2, 5),
(3, 5),\\&
(4, 5),
(5, 5),
(2, 8),
(4, 6),
(3, 7).
\end{align*}
Overall the dataset contains 130k question-answer pairs.

\subsubsection{Model Training}
We use a simple off-the-shelf training setup using the HuggingFace \texttt{transformers v4.30.2} library with a pre-trained T5 model~\citep{raffelExploringLimitsTransfer2020}, which is designed for sequence-to-sequence translation. We use the \texttt{t5-base} version of the model, with 220 million parameters.

For fine-tuning on our task, we hold out 25\% of the dataset that is used in the main text experiments for calibration and testing. The remaining 75\% of the data is used for model training and validation. To this end, we further split the training data into 75\% training data and 25\% validation data. The data is tokenized and encoded using the default \texttt{T5Tokenizer} of \texttt{transformers}.

The model is used with the standard training pipeline of the Transformers library with Pytorch models, defined through the \texttt{Seq2SeqTrainer} interface. We use a learning rate of $2.0\times 10^{-4}$, weight decay of $0.01$ and dropout rate of $0.1$ and train over 15 epochs using the default optimizer of the HuggingFace implementation. A random seed was chosen and fixed for reproducibility. The model was not aggressively tuned as optimizing performance was not the goal. We will release the training code as well as the trained weights.

\subsection{Chemical reactions task}
\subsubsection{Dataset}
We use 479,035 chemical reactions from the SPTO-MIT dataset~\citep{jin_predicting_2017}, which specifies reagents and product molecules with the SMILES scheme for specifying chemical structures~\citep{weiningerSMILESChemicalLanguage1988}. The dataset is used to specify a sequence-to-sequence prediction task: given a list of reagent molecules, predict the product molecule.

SMILES strings are tokenized and encoded using the scheme of~\citet{schwallerMolecularTransformerModel2019} which isolates atom and ion specifiers and structural components.

\subsubsection{Model Training}
Given the similar structure, we also train a T5 model for this task, but from scratch in this case and using the HuggingFace \texttt{t5-small} architecture, which we empirically found to work best on the validation data.

For training, start by splitting the entire dataset into 40k held-out samples used for CP calibration and testing, leaving 439k training samples, which we further split into 30k validation example and 419k samples used properly for training.

We use the \texttt{NanoT5} training pipeline~\citep{nawrot2023} for faster training. We used the \texttt{AdamWscale} optimizer from~\citet{nawrot2023} with a learning rate of $5\times 10^{-4}$, a cosine scheduler with a warmup of 7k steps and a final learning rate of $5\times 10^{-5}$, trained over 12 epochs with a batch size of 16. We used a weight decay parameter of $0.01$ and a dropout rate of $0.1$. These parameters were found starting from the recommendations in~\citep{nawrot2023} and doing manual adjustments based on validation performance until a reasonable performance was reached. Again, obtaining a reasonable performance was enough for this study as reaching or beating state of the art for the task was not our goal.

\section{Software and Hardware details}
All our experiments are run in a Linux HPC cluster. For model training and inference, we used 32-core, 64GB RAM environments with a single NVidia A100 GPU. Experiments were run in Python 3.10 with \texttt{pytorch v2.0.1+cu117} and \texttt{transformers v4.30.2}.

\section{Theoretical tools}
In this section, we cite the theorems that support the proofs in the main text.

\subsection{Conformal Beam Subsets: Proposition 1}
Proposition 1 relies on the conditional decomposition of the sub-beam coverage probability:
\begin{multline*}
    \prob{S\in C(X)} = \prob{S\in \beta(X)} \\
    \times \prob{S \in C(X) | S\in \beta(X)},
\end{multline*}
Where $C(X)$ is the conformal subset of the beam search result $\beta(X)$ we define in section 4.1.

The second term is bounded with the group-conditional conformal guarantee of~\citet{vovkConditionalValidityInductive2012} (Proposition 3.), which allows conditional conformal predictions per group, despite the per-group calibration sample size being a random variable.
\begin{proposition}[\Citet{vovkConditionalValidityInductive2012} (Proposition 3.)]
    Let $\set{(X_i,y_i,g_i)}, i=1,\dots,N$ be an exchangeable calibration set with a group parameter $g_i$. Performing conformal calibration on each group separately yields the following conditional guarantee for a text point $(X,y,g)$ exchangeable with the calibration data:
    \begin{equation*}
        \prob{y\in C_\alpha(X) | g=g_0}\geq 1-\alpha.
    \end{equation*}
\end{proposition}
In our case the group index is the boolean variable $S\in \beta(X)$. This is a peculiar use of conditional conformal coverage given that the second group has 0 coverage: $\prob{S\in C(X) | S\not \in \beta(X) } = 0$.

The first term is bounded \textit{post-hoc} with a binomial confidence interval, given that the calibration beam coverage follows a binomial distribution. Following~\citet{clopperuseconfidencefiducial1934},
\begin{theorem}[Clopper-Pearson Confidence Interval]
    let $X$ be a binomial random variable with parameters $p,n$. Define the random variable $p_\text{min} = \func{B}{\delta; X, n+1-X}$, where $\func{B}{\delta; a, b}$ is the $\delta$-quantile of the beta distribution $\func{B}{a,b}$. Then
    \begin{equation*}
        \prob{p_\text{min}\geq p} \geq 1-\delta.
    \end{equation*}
\end{theorem}

\subsection{Dynamic Conformal Beams: Proposition 3}
Our dynamic conformal beam decoding scheme is an iterative set prediction procedure, where at each step we keep sequences that have truncated scores above a certain thresholds for all previous truncations. At calibration time, we define these thresholds by considering calibration sequences at step $k$ that have passed the thresholds at ealier steps. This calibration procedure follows the multivariate confidence region definition of~\citet{tukeyNonParametricEstimationII1947}, which extends those of \citet{waldExtensionWilksMethod1943}:

\begin{proposition}[Multivariate Confidence Regions of~\citet{tukeyNonParametricEstimationII1947}]
    Let $W_0=\set{X_1, \dots, X_N}$ be exchangeable random variables, $\phi_1, \dots, \phi_p$ be real-valued functions on $X$ and $k_1,\dots, k_p$ integers such that $\sum k_i\leq N$.
    Iteratively, over $j=1,\dots, p$, define
    \begin{itemize}
        \item $f_j$ as the $k_j$-th smallest value $\phi_j(X_i)$ for $X_i\in W_{j-1}$.
        \item $W_j$ as $W_{j-1}$ from which the $k_j$ smallest $\phi_j(X_i)$ values were dropped.
    \end{itemize}

    Then, if the distributions of $\phi_i(X)$ are continuous, a new sample $X$ exchangeable with the samples in $W_0$ verifies
    \begin{equation*}
        \prob{\set{\phi_i(X) \geq f_i}_{1\leq i\leq p}} \sim \func{B}{N+1, N+1-K},
    \end{equation*}
    where $B$ is a beta distribution and $K=k_1+\dots+k_p$.
\end{proposition}

Our calibration procedure reproduces exactly these steps using the length-normalized conditional score on truncated sequences $\func{\pi}{S_{|k}|X}/|S_{|k|}|$ where $|S_{|k|}|$ is the length of the $k$-truncated sequence $S_{|k|}$, ignoring potential padding tokens. At decoding time, we generate all sequences that have a score above this limit conditioned on the input so that if the true sequence is in the confidence region, it is included in the prediction set.

\end{document}